# Interpreting Hand gestures using Object Detection and Digits Classification

**Mrs. Sangeetha K[1], Balaji VS[2], Kamalesh P[3], Anirudh Ganapathy PS[4]**

1Assistant Professor, Department of Artificial Intelligence and Machine Learning,

Rajalakshmi Engineering College, Thandalam, Chennai-602105

sangeetha.k@rajalakshmi.edu.in

2III Year AIML, Department of Artificial Intelligence and Machine Learning,

Rajalakshmi Engineering College, Thandalam, Chennai-602105

211501017@rajalakshmi.edu.in

3III Year AIML, Department of Artificial Intelligence and Machine Learning,

Rajalakshmi Engineering College, Thandalam, Chennai-602105

211501039@rajalakshmi.edu.in

4III Year AIML, Department of Artificial Intelligence and Machine Learning,

Rajalakshmi Engineering College, Thandalam, Chennai-602105

211501009@rajalakshmi.edu.in

**Abstract**

Hand gestures have evolved into a natural and intuitive means of engaging with technology. The objective of this research is to develop a robust system that can accurately recognize and classify hand gestures representing numbers. The proposed approach involves collecting a dataset of hand gesture images, preprocessing and enhancing the images, extracting relevant features, and training a machine learning model. The advancement of computer vision technology and object detection techniques, in conjunction with OpenCV's capability to analyze and comprehend hand gestures, presents a chance to transform the identification of numerical digits and its potential applications. The advancement of computer vision technology and object identification technologies, along with OpenCV's capacity to analyze and interpret hand gestures, has the potential to revolutionize human interaction, boosting people's access to information, education, and employment opportunities.

<u>Keywords:</u> Machine learning, Deep Learning, Neural Networks, Open CV

# 1 Introduction

The classification of hand gestures as numbers is an emerging and significant research area that involves the development of algorithms and techniques to recognize and interpret hand movements that correspond to numerical digits. Hand gestures have always been an integral part of human communication and are widely used in various contexts, including sign language, sports, and everyday interactions. Leveraging machine learning and computer vision methods to accurately classify hand gestures as numbers holds great potential for revolutionizing human-computer interaction, accessibility and accurately classifying hand gestures as numbers holds great potential for revolutionizing human-computer interaction, accessibility, and a wide range of applications. The ability to accurately recognize and classify hand gestures as numbers can have profound implications in numerous domains. In the realm of human-computer interaction, it can provide a more intuitive and natural way for users to interact with digital devices. This could eliminate the need for traditional input methods such as keyboards or touchscreens, offering a more immersive and seamless user experience. Furthermore, for individuals with physical disabilities or conditions that limit their ability to use conventional input devices, hand gesture recognition can open up new possibilities for communication and control. Object Detection is utilized to locate and identify objects based on their respective classes while also determining their precise positions.[1] There are two potential approaches: The first approach involves using a glove-based system where the user wears specialized gloves that record their hand movements. The second approach focuses on vision, which can be categorized into static and dynamic recognition. Dynamic recognition concentrates on real-time gesture tracking, whereas static recognition involves representing motions in a two-dimensional space.[3] Real-time hand gesture detection finds its primary application in recognizing and interpreting hand gestures. Various algorithms, including artificial neural networks and computer vision techniques, are employed to comprehend different hand gestures.

# 2 Related Work

CNN, or Convolutional Neural Network, is a type of artificial neural network that operates through the adjustment of weights. In CNN, convolution is the result of merging functions that demonstrate the overlap of functions with another. To acquire maps, the snapshot is combined with the activation method that triggers the output. Subsequently, these maps are processed using pooling layers to create conceptual feature maps that reduce the intricacy and difficulty of the network. By iteratively applying the desired quantity of filters, feature maps are produced. The maps involved in fully connected layers generate the output for image detection and exhibit confidence scores for class labels.

The CNN operational model is shown below.

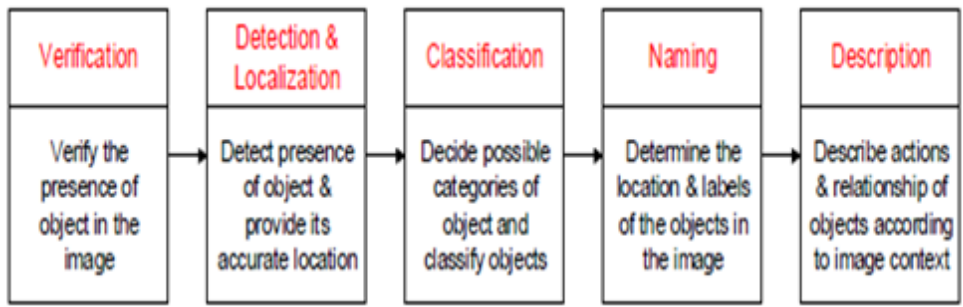

[3]The collection of imageset consisted of approximately 216 images, which were subsequently partitioned into 5 distinct sets. Each set contained an average of 42 images. The process of assigning labels involved employing the You Only Look Once (YOLO) format, where in particulars such as the identification of the class and its corresponding class were stored. The dataset was labeled in adherence to the YOLO format and saved in a text file format. The dataset encompassed a total of 216 images, distributed across 5 different sets denoted as set-1, set-2, set-3, set-4, and set-5, which depicted various finger positions. The hand gestures were visually represented in the dataset.

The hand motions in our dataset are shown in Figure 1.

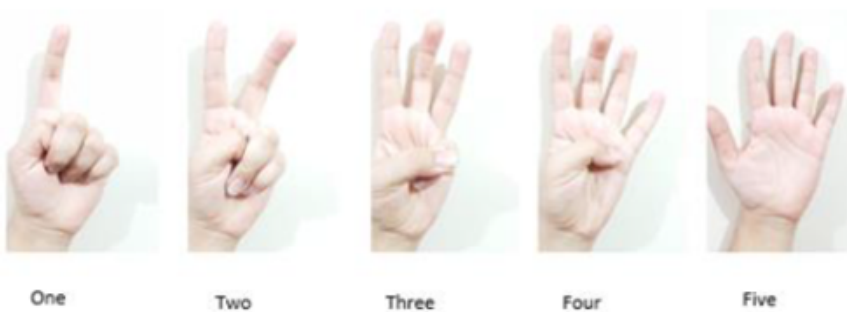

Figure 1. Hand gesture set to be detected.

In [3], utilizing the OPENCV library, a freely available computer vision tool is employed to capture objects in real-time, and subsequently, images are acquired and processed in conjunction with these real-time objects. The dataset includes class annotations associated with the images, as they are annotated based on the classes specified for the YOLO technique. Consequently, the training phase can be initiated. When input is received, it enters the input layer for initial processing before passing through interconnected convolutional hidden layers, which consist of multiple convolutions with varying sizes. The convolutional layers contribute to determining a distinct confidence value.

[2]This approach involves utilizing Tensorflow, a computer vision methodology, for the pixel-wise segmentation of images. The implementation can be carried out in four distinct stages: image segmentation and enhancement, orientation detection, feature extraction, and categorization. However, one significant limitation of this approach, which revolves around the aforementioned four stages, is the challenge posed by diverse lighting conditions that rapidly alter color properties, potentially resulting in errors or failures. For instance, inadequate lighting may render the hand region invisible, while non-skin areas with similar color properties may be misidentified as the hand region. The system is designed to model hand gestures in the spatial domain, employing various two-dimensional and three-dimensional functional and non-functional models for simulation. Nevertheless, the primary drawback is the system's failure to consider temporal gestures or fist movements, and its limited ability to detect images with complex backgrounds or the presence of new objects alongside fist-related objects within the frame. This study primarily focuses on comprehending motions through diverse data collection, initial processing, categorization, and other methodologies. Suitable tools for data collection, such as information gloves, markers, and palm images, should be employed. However, it is essential to acknowledge the limitations of the system in terms of image clarity, rotation, direction, scale, and specifications.

[4] Within a computer vision undertaking employing OpenCV and Python, the process of gathering data encompasses multiple stages that aim to ready the dataset for training a model capable of detecting objects. To commence, the project makes use of the "os" module to manage file paths, enabling streamlined navigation and manipulation of image and XML files. Moreover, the "time" module is employed to assess code efficiency and track the duration of various operations. The "uuid" library assumes a crucial role in generating unique identifiers for image files, ensuring their distinctiveness and mitigating the possibility of conflicts during data processing. The dataset is annotated utilizing the LabelImg tool, a widely adopted annotation utility for computer vision projects. This tool empowers users to label objects of interest within images, generating XML files that encompass comprehensive information pertaining to the labeled objects, including their coordinates and corresponding class labels. Following the annotation process and XML file creation, the data collection is partitioned into training and validation subsets, often adhering to an 80-20 distribution ratio. This partitioning scheme ensures that the system is trained on a significant portion of the data, while preserving a separate set of images for evaluation and performance assessment purposes. In order to streamline the training

procedure, a labeled map is established, associating each letter of the alphabet with a unique numerical ID ranging from one to twenty-six. This map acts as a reference for interpreting and mapping the data labels assigned to objects during the annotation phase. To build, train, and utilize the provided model, the project harnesses the TensorFlow object detection API, which offers a comprehensive framework for constructing robust object detection models. The TensorFlow detection model zoo encompasses a diverse array of pre-trained models that serve as effective starting points for the project. Leveraging these pre-trained models facilitates transfer learning, allowing the model to leverage existing knowledge and adapt to specific object detection tasks. The data acquisition process in a computer vision project employing OpenCV and Python involves utilizing the os, time, and uuid modules, annotating images with the LabelImg tool, partitioning the dataset, creating a labeled map, and harnessing the TensorFlow object detection API with pre-trained models. These steps collectively ensure a systematic and efficient approach to acquiring, preparing, and utilizing the dataset for training an object detection model.Gesture recognition holds significant importance in the field of human-computer interface technology. Its applications range from audio recording and automated system monitoring to visual communication and graphical surveillance.

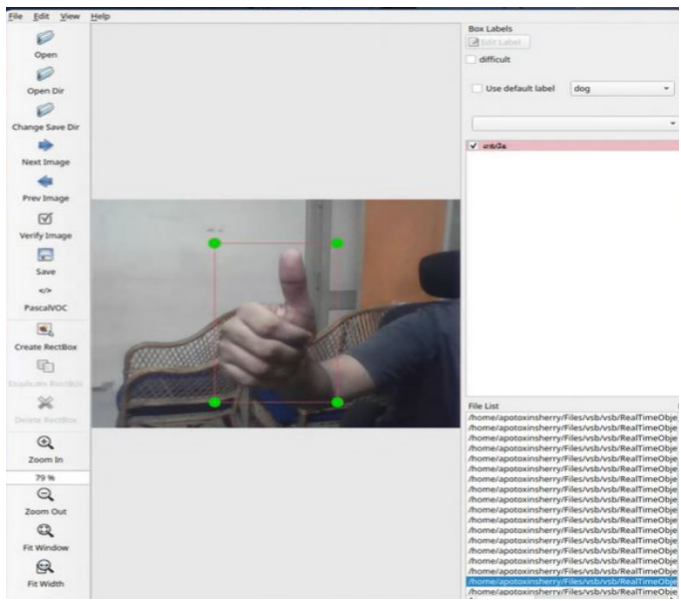

Fig 2: Labeling images using labelimg

The system proposed in [5] aims to develop a hand gesture detection system using MediaPipe's pre-trained model, which demonstrates real-time detection and tracking of hands in video streams. To achieve this, the neural network model will be trained and fine-tuned using the open-source machine learning framework, TensorFlow, leveraging MediaPipe input to detect and categorize various hand gestures. Furthermore, the advanced computer vision library, OpenCV, will be utilized to analyze video frames, extract crucial hand features, and preprocess the data for input to the neural network.

By employing OpenCV's extensive collection of image processing and computer vision operations, the accuracy and robustness of the hand gesture detection system will be enhanced. The dataset employed for training consisted of approximately 216 images, which were divided into five distinct groups, each containing an average of 42 images. The labeling process followed the You Only Look Once (YOLO) format, capturing details such as the class ID and the corresponding class information, which were stored for reference. The dataset was labeled in the YOLO format and saved as a text file. It comprised 216 images, categorized into five separate sets, namely set-1, set-2, set-3, set-4, and set-5, representing different finger positions. The dataset encompassed visual representations of various hand movements.

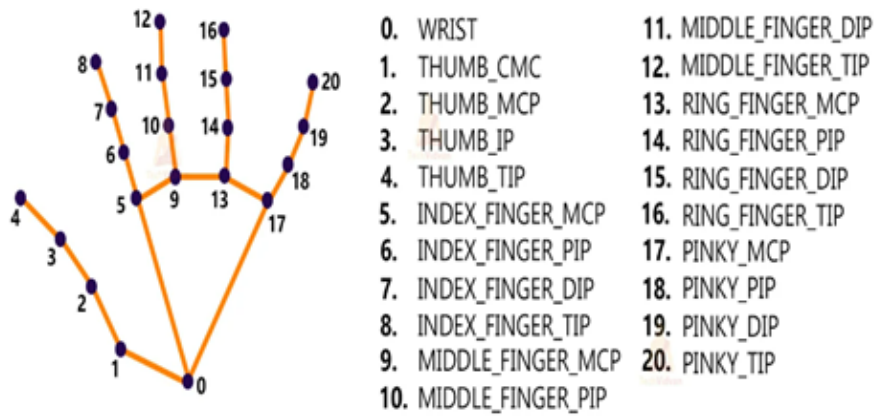

Fig 3 : 21 points the hand

[9]A unique real-time approach for hand gesture identification that employs an appearance model has been created. Most hand detection systems are hampered by the presence of complicated backgrounds. Existing skin color-based techniques are unreliable because separating fingers among things that have comparable skin tones is challenging, especially in low-light circumstances. To be successful, shape models require a noticeable difference across their fingers and the background. The Adaboost technique algorithm has been used in attempts to recognise hands in grayscale photographs.

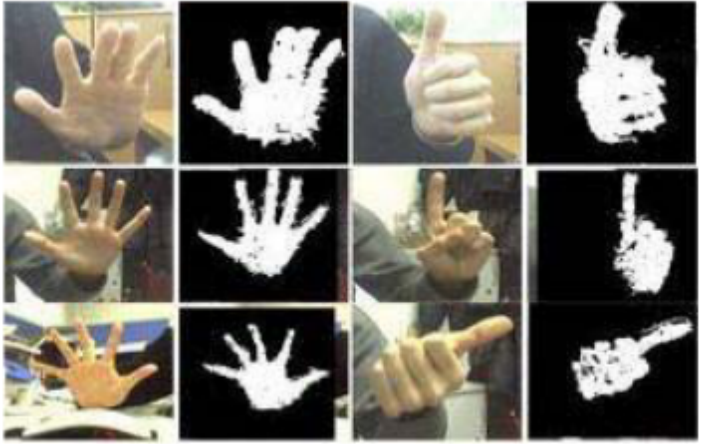

Fig 4 : Hand-segmentation results

Hand detection is critical in the early stages of human-computer interaction, particularly with gesture interfaces, where it functions as a trigger for turning on the control panel. After detecting a hand, we collect hue information from the surrounding region near the average position of the observed features. To achieve a compromise between computing economy and precision we use just one Gaussian structure to characterize the color of the hand in the HSV hue space throughout the picture identification operation. Object tracking, such as tracking hands, is a substantial difficulty. While shape-based approaches produce good results for stiff objects, they are unsuitable for articulated items such as hands. Texture or appearance-based approaches have been enhanced to increase their resilience for non-rigid objects. Some methods include backdrop modeling, however this is restricted to the arrangements with immovable cams. When the object has only minor deformations, optical flow-based approaches can produce good tracking results.

[10] For recognising gestures for controlling television, Freeman and Weissman created a model based on normalized correlation. The photographs are recorded and saved in the database. The database is made up of thirty photos that may be interacted with. The color of the photos is then altered from their original colour to ycbcr to eliminate the shadow issues. The skin's possibility is defined by the min and max values. Furthermore, the noise in the photos is eliminated. Noise can be present both around and within the hand.Fig 4. The image's characteristics and properties are then extracted using feature extraction.

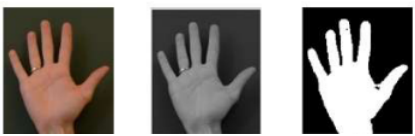

Fig 5: The different color of the images of the hand

The hand is then detected correctly and precisely by drawing a box around it (see Fig 5).The box is known as the bounding region, and it is used to accurately recognise the motion.

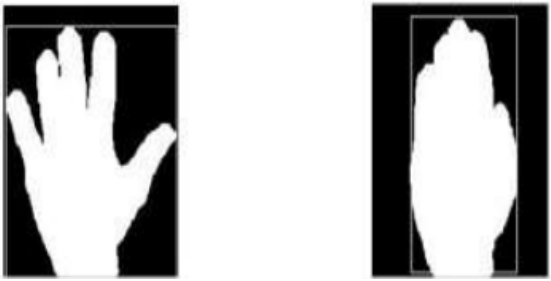

Fig 6: Bounding region in hands

[12]This paper proposes a novel approach to action recognition that involves integrating data and picking the most important attributes. In the beginning, an HSI colour transformation is used to increase the contrast of video frames. The next step is to extract motion data from moving regions using an optical flow technique. These parts are then combined with form and texture components .On the combined vector, a novel Weighted Entropy-Variances technique is used to identify the most optimum attributes for classification. SVM is then employed to group human actions after receiving the specified features as input.

To classify the gestures, the technique uses neural networks composed of convolutions using raw input from sensors. The main challenge is determining the best CNN architecture. To address this, we must first define the system's input-output structures. We propose the optimum design determined by the results of various tests. Every captured instance is made up of a (eightXeightXN) construction, with the height (N) of the framework altering depending on how the subjects move individually.

A specified dimension for input data is essential for successful implementation. One method is to base the given size on the group with the highest framework height and then apply lowest padding for situations with shorter frames. Nonetheless, this approach creates a large input array, which incurs significant processing costs. To split the findings into fixed-length subsamples, we use a different approach. This entails breaking down each particle into smaller bits.

Advantages of this approach are:

(i) Since the duration of the frame increases, so does the amount of samples required to develop the technique of the simulated network. More sections with less details are provided by shorter frame lengths, and vice versa.

(ii) By constructing pieces from various elements of the initial collection, we may infer the type of interaction regardless of whether it was not clearly mentioned at the outset.

(iii) Unlike other systems for classifying following motions, our system classifies actual time. After obtaining a predefined amount of data, it identifies the action. The output version of our approach adopts a soft-max mechanism with fourteen classifications.

[13] The article proposes an innovative structure for categorizing items in a system that varies from previous multifaceted deep learning techniques by significantly reducing parameters and shortening inference time. As a result, this characteristic is particularly suited to real-time applications. In essence, the decreased network affects classification efficiency somewhat. The analysis, on the other hand, demonstrates that this model strikes a good balance between speed and clarity, making it a great candidate for developing an autonomous machine capable of item classification and location extraction.

[14]We describe a unique strategy for using Convolutional Neural Networks, to create somewhat weaker classifiers. By combining these models, we create a more powerful and differentiated model. We use (MV-CNN) technique to handle visual data, which employs many projection views of a 3D model to improve efficiency over individual projections. We adapt neural network topologies for voxel information and train them on different orientations of each item in the training dataset. This allows us to store features other than those obtained from pixel data.The Volumetric CNN has three 3-dimensional layers of convolution and two fully linked layers, with the final layer acting as a classification system. These layers' convolutional kernels are intended to record voxel relationships throughout the object's depth. We may comprehend vast spatial linkages inside the object along all axes by training the network on numerous orientations of all models. sparse locally connected kernels is used to minimize computational cost, using the 3D nature of voxels to overcome spatial adaptation limits imposed by two-dimensional graphical CNNs due to kernel size constraints.The ReLU layer, which comes after the convolutional layer, introduces non-linearity, which is critical for successful class distinction. Furthermore, the succeeding layer of pooling minimizes duplicate data collecting, reducing model size. We use 3x3 kernels in our Volumetric Convolutional Neural Network because they are sufficient for recognising associations in voxelized data, as seen by the obvious demarcation of separate areas measuring 30x30. Leavers are employed to avoid overfitting.

We use a completely connected layer with forty neurons as the classifier for the ModelNet40 dataset, and a network that is completely connected with ten neurons as the classifier for the ModelNet10 dataset. The FusionNet approach combines many networks in the last fully linked layer, resulting in improved classification accuracy when compared to each separate network component.

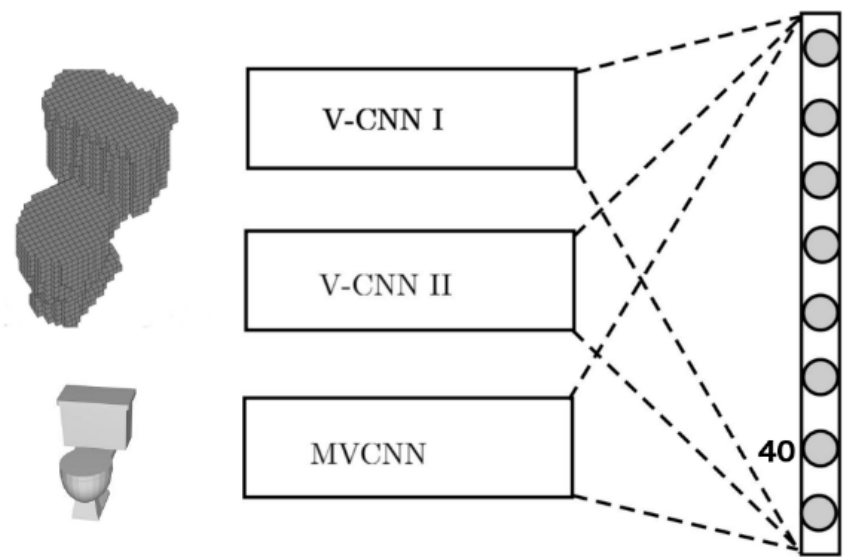

Fig 7: Combining two CNNs

[15] The method has its foundation on DNN. It is made up of seven layers, two of which are linked and the rest are convolutional layers. For the non-linear conversion process, each layer must employ Relu. Convolution layers are important in the max pooling process.The last layer constitutes a regression layer that creates an Object-binary-mask with a focussed dimension output. The value of class can be either 1 or 0, with 1 indicating that the pixel is within the bounding box of class and 0 indicating that it is outside. Because of its non-convex nature, the base-network's optimality cannot be guaranteed. Furthermore, regularization of the loss function is required by tweaking and altering the weights that are used for each generated output. To address this issue, outputs that are much less than the picture size may be assigned a value of zero. To remedy this, the weights are changed and raised for non-zero values using the lambda parameter. When lambda is small, values that are less than one are penalized, causing the architecture to anticipate values other than zero even in the presence of poor signals.

[16] We developed a revolutionary strategy that integrates object identification and hybrid artificial neural network approaches into a single system. Our technique merely forecasts single bounding boxes but additionally determines limits for every category present in the picture by taking image-wide factors into account. The network can completely analyze both the entire shot .We achieve extensive training while maintaining accuracy and assuring actual time analytical capabilities by leveraging the YOLO system. To do this, we divide the input picture into a grid of SxS cells, with each cell responsible for recognising an item if its center sits inside the borders of that cell. We construct projections for the box's perimeter (B) and its related grid cells within every single cell.

## 3 Implementation

Hand gestures are identified and converted into the local language using a series of steps. The process begins with the creation of a set of pre-labeled images through labeling. Each gesture is represented by around 20 images, and these images are annotated with corresponding labels. This labeling step is crucial for real-time detection, as it assigns specific labels to each gesture.Next, transfer learning is employed by utilizing the SSD MobileNet model.Transfer learning is a machine learning technique where a model is trained on a large dataset for a specific purpose and then repurposed or modified for a different task. By employing transfer learning with the SSD MobileNet architecture, a pre-trained model for image classification and object detection is utilized. MobileNet is a well-known deep learning framework used for image classification, while SSD (Single Shot Detector) is a framework used for object detection.The third step involves real-time detection using a webcam and the OpenCV library for object detection. Once the training and labeling of the images are complete, the model is ready for real-time detection. The system is expected to achieve high accuracy in detecting and classifying hand gestures, providing reliable results.

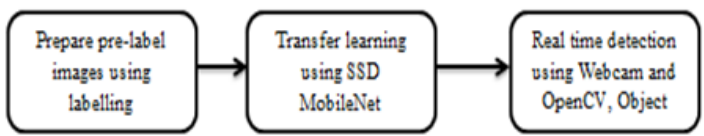

Fig 8: Demonstration of the model

The images that have undergone training are subjected to preprocessing to extract relevant information related to the hand gestures. Subsequently, features are extracted from the provided set of trained gestures and stored in individual files for each gesture. These files contain the extracted characteristics of the gestures and serve as a reference for future testing.The testing phase follows a similar process, wherein images are subjected to the same flow. The images are preprocessed and features are extracted from them using the previously trained model. These features are then compared to the stored features to identify the closest match or similarity with a known gesture. This enables the real-time detection of gestures. Once a gesture is detected, the associated text is recognized and predicted based on the recognized gesture.

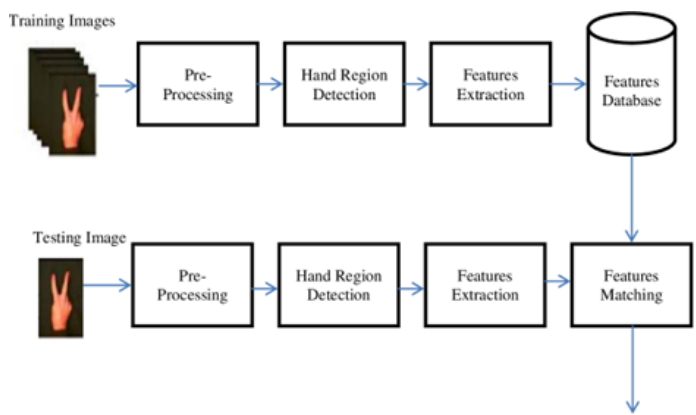

Fig 9: Steps in training and testing.

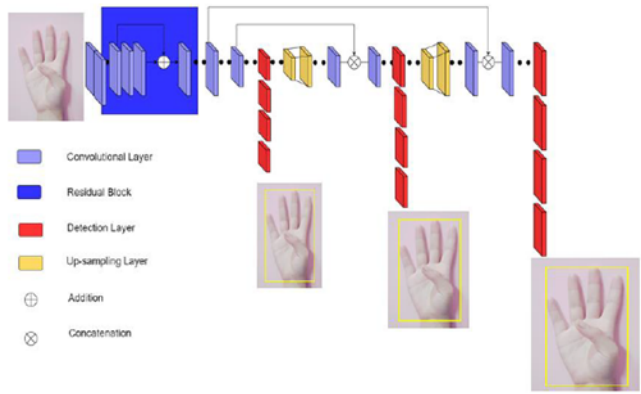

Fig 10: Hand gesture recognition using neural networks

The process begins with an input being received by the input layer. This input is initially displayed and then forwarded to the hidden layers, which consist of multiple convolutional layers with varying sizes and quantities. These convolutional layers collectively contribute to determining a unique confidence value.For evaluating the model's performance, a testing sequence of 10,000 steps was defined, utilizing the same hyperparameters employed during training. Throughout the training process, the model experiences losses in categorization, regularization, and localization. The localization loss quantifies the difference between the predicted and actual improvements in bounding boxes. This loss varies depending on the modifications made to the displayed bounding boxes and their true values. Following training, the model is populated starting from the most recent checkpoint to enable real-time detection. Configuration and updates are performed to prepare the model for instruction. The instructed model is then loaded with the latest checkpoint, finalizing the preparation process for real-time sign language recognition. This entire process can be further adapted to meet specific requirements, including those related to local language integration.

During the implementation time, we have received the following accuracy and loss.

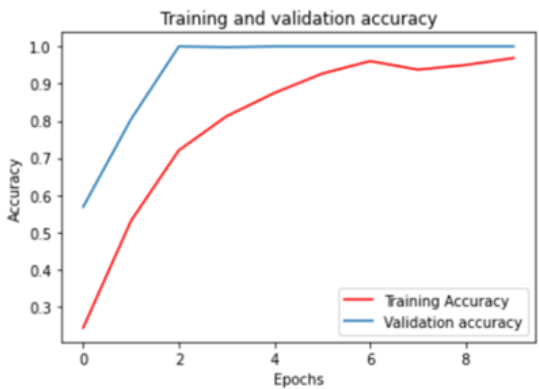

Fig 11: Training vs Validation accuracy

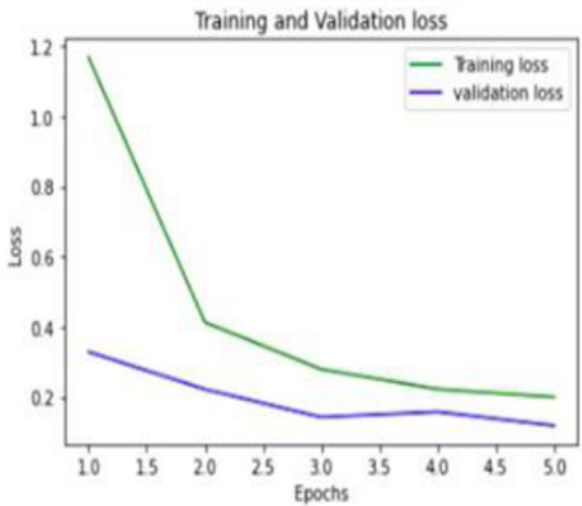

Fig 12: Training vs Validation loss

## 4 Results and discussions

The system performs effectively and precisely recognizes the hand gesture. Subsequently, each movement is assigned a corresponding label and translated into the local language to enhance comprehension. For a specific gesture, a dedicated dataset comprising 13 images is assembled, stored in an independent directory, labeled with the respective gesture name, and subsequently employed for real-time recognition. By collecting multiple images of each gesture and training the model, the losses can be computed and minimized, enhancing the system's performance and accuracy.

## 5 Conclusion and Future Works

The implemented model demonstrated robust performance and achieved a high level of accuracy in real-time gesture recognition and classifying it to digits. By assigning appropriate labels ,the system enhances usability and facilitates effective communication. There are several avenues for future research and improvement. Firstly, expanding the dataset by incorporating a wider range of hand gestures can enhance the model's capability to recognize diverse gestures accurately. Secondly, exploring advanced training techniques, such as transfer learning or data augmentation, could further enhance the model's performance and generalization ability.

Additionally, investigating different neural network architectures or leveraging more advanced algorithms may yield improved results. Moreover, integrating the hand gesture recognition system with other applications or devices, such as augmented reality interfaces or smart home systems, can extend its utility and enable innovative interactions. Further research can also focus on real-time gesture recognition in dynamic environments and addressing challenges related to occlusion, varying lighting conditions, and complex backgrounds. Overall, this project lays the groundwork for future advancements in hand gesture recognition technology and opens up opportunities for its integration into numerous domains, including human-computer interaction, assistive technologies, and virtual reality systems.
.